\documentclass[lettersize,journal]{IEEEtran}
\usepackage{amsmath,amsfonts}
\usepackage{algorithmic}
\usepackage{array}
\usepackage[caption=false,font=normalsize,labelfont=sf,textfont=sf]{subfig}
\usepackage{textcomp}
\usepackage{stfloats}
\usepackage{url}
\usepackage{verbatim}
\usepackage{graphicx}
\usepackage{booktabs}
\usepackage[table]{xcolor}  

\def\BibTeX{{\rm B\kern-.05em{\sc i\kern-.025em b}\kern-.08em
    T\kern-.1667em\lower.7ex\hbox{E}\kern-.125emX}}
\usepackage{balance}

\begin{document}
\title{A Layered Self-Supervised Knowledge Distillation Framework for Efficient Multimodal Learning on the Edge}
\author{Tarique Dahri $^{1}$, Zulfiqar Ali Memon $^{1}$, Zhenyu Yu $^{2,}\IEEEauthorrefmark{1}$, Mohd. Yamani Idna Idris $^{2}$, Sheheryar Khan $^{3}$, Sadiq Ahmad $^{4, }\IEEEauthorrefmark{2}$, Maged Shoman $^{5}$, Saddam Aziz $^{6}$ and Rizwan Qureshi $^{7,}\IEEEauthorrefmark{1}$,~\IEEEmembership{Senior Member,~IEEE}

\thanks{
    $^{1}$ Fast School of Computing, National University of Computer and Emerging Sciences, Karachi, Pakistan;\\
    $^{2}$ Universiti Malaya, 50603 Kuala Lumpur, Malaysia;\\ 
    $^{3}$ School of Professional Education and Executive Development, The Hong Kong Polytechnic University, Hong Kong;\\ 
    $^{4}$ COMSATS University Islamabad, Wah Campus, 47040, Wah Cantt, Pakistan;\\ 
    $^{5}$ Intelligent Transportation Systems University of Tennessee-Oak Ridge Innovation Institute’s Energy Storage and Transportation Convergent Research Initiative;\\ 
    $^{6}$ Independent Researcher, USA; \\
    $^{7}$ Center for research in Computer Vision, University of Central Florida; Orlando, Florida, USA;\\ 
    \IEEEauthorrefmark{2} These authors also contributed equally to this work. \\
    \IEEEauthorrefmark{1} Correspondence: Zhenyu Yu (yuzhenyuyxl@foxmail.com); \\Rizwan Qureshi (engr.rizwanqureshi786@gmail.com) \\
    }
}



\maketitle

\begin{abstract}
We introduce Layered Self-Supervised Knowledge Distillation (LSSKD) framework for training compact deep learning models. Unlike traditional methods that rely on pre-trained teacher networks, our approach appends auxiliary classifiers to intermediate feature maps, generating diverse self-supervised knowledge and enabling one-to-one transfer across different network stages. Our method achieves an average improvement of 4.54\% over the state-of-the-art PS-KD method and a 1.14\% gain over SSKD on CIFAR-100, with a 0.32\% improvement on ImageNet compared to HASSKD. Experiments on Tiny ImageNet and CIFAR-100 under few-shot learning scenarios also achieve state-of-the-art results. These findings demonstrate the effectiveness of our approach in enhancing model generalization and performance without the need for large over-parameterized teacher networks. Importantly, at the inference stage, all auxiliary classifiers can be removed, yielding no extra computational cost. This makes our model suitable for deploying small language models on affordable low-computing devices. Owing to its lightweight design and adaptability, our framework is particularly suitable for multimodal sensing and cyber-physical environments that require efficient and responsive inference. LSSKD facilitates the development of intelligent agents capable of learning from limited sensory data under weak supervision.
\end{abstract}

\begin{IEEEkeywords}
Self-Supervised Learning, Knowledge Distillation, Edge Computing, Multi-modal Learning, Lightweight Deep Models
\end{IEEEkeywords}

\begin{figure}[!ht]
    \centering
    \includegraphics[width=1\linewidth]{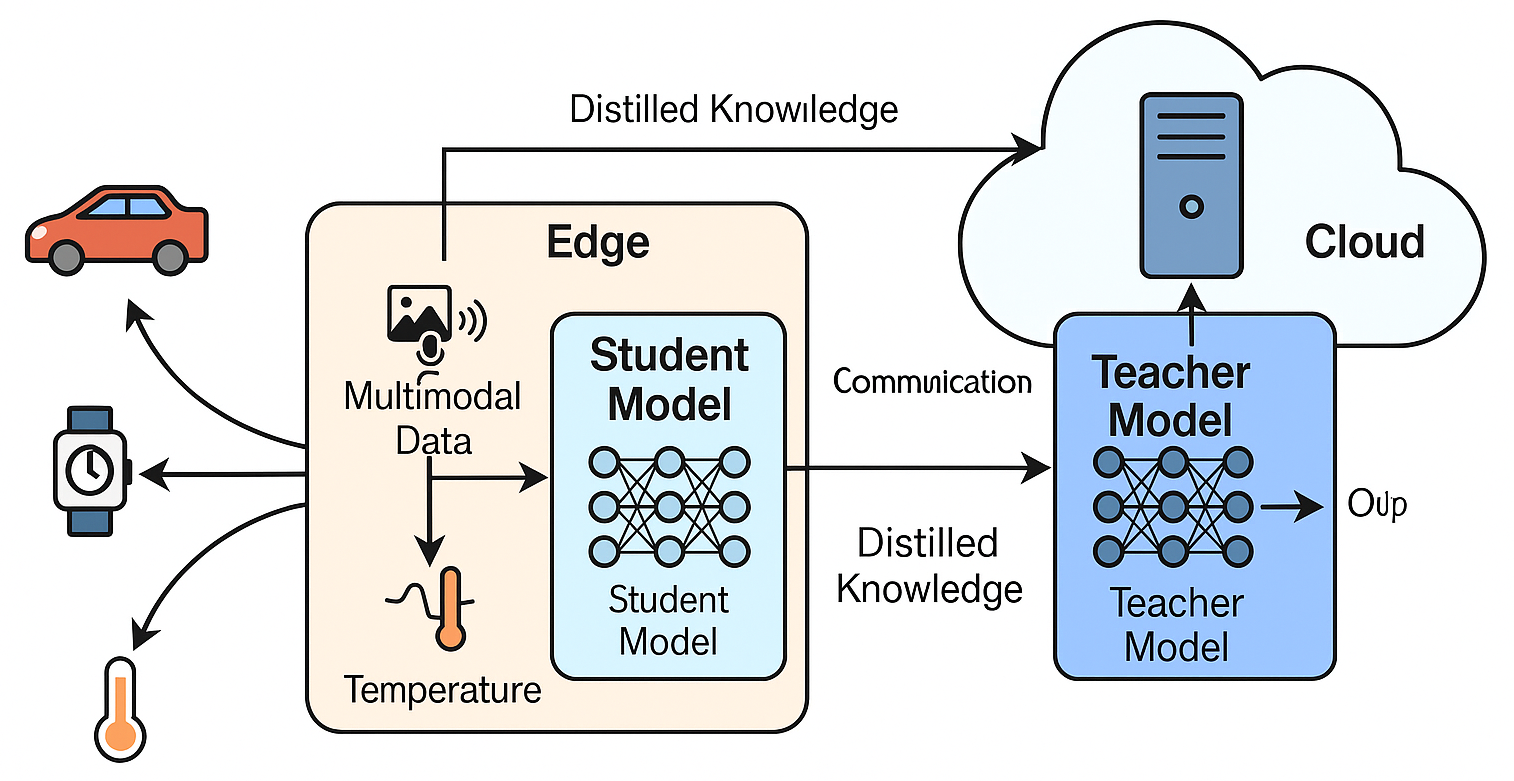}
    \caption{System overview of the proposed Layered Self-Supervised Knowledge Distillation (LSSKD) framework for IoT scenarios. The resource-intensive teacher model is deployed in the cloud, while the lightweight student model runs on edge devices to enable efficient and low-latency inference.}
    \label{fig:framework_overview}
\end{figure}

\section{Introduction}

Deep learning has revolutionized computer vision, achieving remarkable success in tasks such as image classification~\cite{mikolajczyk2018data,yu2024capan,wang2024multi}, object detection~\cite{zhao2019object}, face recognition~\cite{huang2022evaluation}, pose estimation~\cite{zhang2019fast}, activity recognition~\cite{liu2021semantics}, and semantic segmentation~\cite{garcia2018survey,luo2022pixel}. However, these advances often come with increased computational complexity~\cite{thompson2020computational}, memory demands~\cite{yu2021compute}, and energy consumption~\cite{strubell2020energy}, which hinder the deployment of large models in resource-constrained environments such as edge devices~\cite{ji2024edge, yin2024dpal}. The proposed framework separates computation between cloud and edge, enabling lightweight inference on-device and full-capacity training remotely (see Figure~\ref{fig:framework_overview}).

To address these limitations, model compression strategies such as pruning~\cite{han2015learning}, quantization~\cite{wu2016quantized}, and knowledge distillation (KD)~\cite{buciluǎ2006model} have been widely explored. Among them, self-knowledge distillation (Self-KD)~\cite{kim2021self} has gained popularity for enabling a model to learn from its own predictions without relying on a large teacher network. This paradigm significantly reduces training costs while retaining high performance, especially when combined with regularization techniques such as $\mathcal{L}_1$/$\mathcal{L}_2$ weight decay~\cite{nowlan2018simplifying}, dropout~\cite{srivastava2014dropout}, batch normalization~\cite{ioffe2015batch}, and advanced data augmentation~\cite{devries2017improved}. Progressive Self-Knowledge Distillation (PS-KD)~\cite{kim2021self} extends this concept by leveraging predictions from previous epochs as soft targets, forming a temporal self-supervision loop. However, PS-KD only utilizes the final layer outputs, neglecting the valuable intermediate representations within the network hierarchy.

Traditional KD methods~\cite{hinton2015distilling, kim2021self, yu2024improved} typically focus on aligning the final output distributions of student and teacher networks by minimizing the Kullback-Leibler (KL) divergence. While effective, such approaches fail to capture the hierarchical knowledge embedded in intermediate layers. Recent extensions have attempted to incorporate intermediate representations~\cite{zagoruyko2016paying}, but they often rely on large pre-trained teacher networks and do not utilize multi-level label softening or cross-stage supervision strategies.

\begin{figure*}[!t]
\centering
\includegraphics[width=\linewidth]{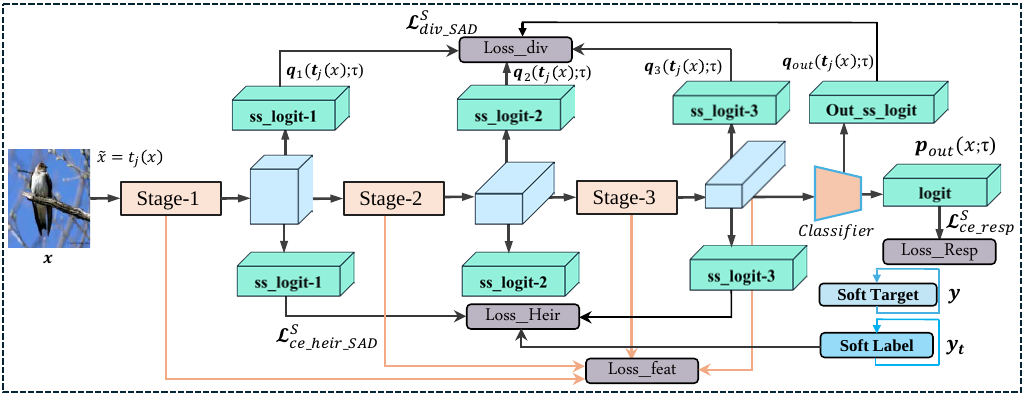}
\caption{Overview of the proposed LSSKD mechanism. Auxiliary classifiers after each bottleneck stage produce self-supervised augmented distributions (SADs), while the final linear classifier provides the predictive distribution. The self-loop denotes progressive softening of hard labels via multi-stage predictions.}
\label{fig_1}
\end{figure*}

To overcome these limitations, we propose the \textbf{Layered Self-Supervised Knowledge Distillation (LSSKD)} framework. As illustrated in Figure~\ref{fig_1}, LSSKD introduces auxiliary classifiers after each bottleneck stage, which generate Self-supervised Augmented Distributions (SADs) to support hierarchical label softening. The final classifier produces the predictive class distribution, while unified soft labels integrate outputs from both shallow and deep stages. Moreover, LSSKD enables knowledge flow from deeper to shallower stages using KL divergence to improve early-layer classifiers. An additional $\mathcal{L}_2$ loss is applied to minimize the discrepancy between auxiliary and final feature maps, promoting internal consistency. Unlike previous approaches such as HCSKD~\cite{yang2021hierarchical} and HASSKD~\cite{yang2022knowledge}, which rely on fixed hard labels, LSSKD leverages conflicting and softened label distributions to enhance learning robustness and generalization (see Figure~\ref{fig_2}). 

Our key \textbf{contributions} are summarized as follows:
\begin{itemize}
    \item \textbf{Hierarchical Label Softening:} We propose a multi-level label softening strategy using probabilistic distributions from intermediate classifiers, enabling gradual refinement of hard targets.
    
    \item \textbf{Cross-layer Distillation:} The framework distills knowledge not only across layers but also across training iterations, using auxiliary self-supervision and feature consistency regularization.
    
    \item \textbf{Empirical Performance:} LSSKD surpasses PS-KD and SSKD by average margins of 4.54\% and 1.14\% respectively on CIFAR-100, and outperforms HASSKD by 1.63\% on average. On ImageNet, it achieves a 0.32\% gain over HSSAKD.
\end{itemize}

\section{Related Works}

\subsection{General Knowledge Distillation}
In knowledge distillation~\cite{hinton2015distilling}, a small student model is typically supervised by a large teacher model or ensemble of models. The concept was first introduced by Buciluǎ et al.~\cite{buciluǎ2006model}, and later refined by Geoffrey Hinton~\cite{hinton2015distilling}, who used the softmax temperature to reduce the probability weights, extracting more useful information from the teacher model. KD has been crucial in addressing the computational and storage demands associated with large deep learning models~\cite{sarfraz2021knowledge}. By distilling knowledge from large teacher models into more efficient student models, KD enables real-time processing and deployment on resource-constrained systems~\cite{doshi2024reffakd}. It compresses knowledge, facilitating efficient utilization of computational resources and deployment on lower-resource devices. The rise of multimodal sensory systems—such as those in autonomous robotics or biomedical sensing—calls for models capable of generalizing across modalities and functioning effectively under data scarcity. LSSKD, by distilling knowledge across hierarchical stages and facilitating efficient representation learning, demonstrates strong potential for deployment in sensory-driven AI systems, where annotated data is limited and system robustness is critical~\cite{yang2025constructing}. Moreover, this approach resonates with the biologically inspired paradigm of layered sensory processing observed in natural intelligence.

Beyond model compression, KD finds applications in various domains, including privileged learning~\cite{lopez2015unifying}, mutual learning~\cite{chen2020online}, assistant teaching~\cite{son2021densely}, continual learning~\cite{li2024continual}, text-image retrieval~\cite{rao2023dynamic}, and others~\cite{gou2021knowledge}. Recently, in Large Language Models (LLMs)~\cite{hadi2023survey}, KD has played a pivotal role in transferring advanced capabilities from leading proprietary LLMs~\cite{xu2024survey}, such as GPT-4~\cite{achiam2023gpt}, to open-source counterparts such as LLaMA-2~\cite{touvron2023llama} and Mistral~\cite{karamcheti2021mistral}.

While knowledge distillation has demonstrated robust performance across various datasets and tasks~\cite{cho2019efficacy}, it has faced challenges achieving similar levels of performance when applied to deeper neural networks, due to issues such as information bottleneck~\cite{kim2019curiosity}, loss of fine-grained details~\cite{hao2022learning}, and overfitting~\cite{chen2020robust}. To enhance accuracy, subsequent works~\cite{ahn2019variational, park2019relational, tung2019similarity} have explored feature-based information to capture representational details hidden in intermediate layers. For instance, Fitnet~\cite{adriana2015fitnets} proposed transferring feature maps between selected pairs of student-teacher layers, attention maps~\cite{zagoruyko2016paying} aimed to mimic the attention map of the teacher network, and Feature Similarity Preservation (FSP)~\cite{yim2017gift} mimicked second-order statistics (Gram Matrix)~\cite{devries2017improved}, enabling the student model to learn relationships between different features. Attention transfer~\cite{heo2019knowledge} transfers information from attention boundaries in the teacher model to the student model. Deep supervision (DS)~\cite{lee2015deeply} was initially introduced to address convergence issues and enhance classification performance by incorporating multiple auxiliary classifiers~\cite{chen2022knowledge} in shallow layers of a deep neural network to facilitate learning of ground-truth labels~\cite{luo2023knowledge}.

\subsection{Self-Supervised Representational Learning}
Recent advancements in KD have integrated representation learning to leverage natural supervision derived from the data. Contrastive Representation Distillation (CRD)~\cite{tian2019contrastive} proposes an objective function measuring mutual information between intermediate representations learned by teacher and student networks. Self-supervised Knowledge Distillation (SSKD)~\cite{liu2021sskd} exploits the similarity between self-supervision signals generated from an auxiliary model for distillation, utilizing the SimCLR~\cite{chen2020simple} framework as a pretext task for natural supervision from data.

Hierarchical Augmented Self-supervised Knowledge Distillation (HASKD)~\cite{yang2021hierarchical} merges representational learning with knowledge distillation by introducing a self-supervised augmented task. This task aims to learn the unified distribution of the classification task and auxiliary self-supervised task. HASKD adopts a hard label-based supervised method to guide the original and auxiliary tasks. While KD suggests soft labels generally generalize better between classes compared to hard targets, HASKD smooths hard targets using past predictions. Following a similar strategy, we continuously refine hard labels using past predictions, leveraging intermediate auxiliary layers to enhance final classifier generalization. This approach effectively exploits intermediate auxiliary layers, improving final classifier generalization. Although prior work on knowledge distillation has focused predominantly on visual tasks and natural language processing, limited attention has been given to sensory-rich systems or multimodal environments. Our method provides a promising foundation for compact inference models in applications such as wearable health sensors, multimodal perception systems, and energy-constrained robotics.

\subsection{Consecutive Self-Knowledge Distillation}
In Self-supervised KD (SSKD)~\cite{yuan2020revisiting, yun2020regularizing}, the student model learns from its own predictions during training, enabling efficient knowledge transfer, enhanced generalization, and reduced complexity~\cite{niu2022scale, xu2020knowledge}. For example, CSKD~\cite{yun2020regularizing} regularizes predictive distribution among samples of the same class. Tf-KD~\cite{yuan2020revisiting}, known as Teacher-free KD, employs self-training and manually-designed regularization. Self-training involves the student model learning from its own predictions, replacing dark knowledge with model predictions. Manually-designed regularization uses a virtual teacher model with 100\% accuracy as a regularization term. Tf-KD achieves comparable performance to normal KD without requiring a stronger teacher model or additional computational cost. PS-KD~\cite{kim2021self} uses past predictions of the last layer for continuous refinement of hard targets. We adopt a similar strategy, using soft targets as regularization~\cite{kim2021self, yuan2020revisiting, you2017learning}. However, our method differs in key aspects. Firstly, we employ self-supervised learning techniques to derive natural supervision directly from the data themselves, capturing intrinsic semantic and pose information throughout the network for improved performance. Additionally, our approach utilizes predictions from intermediate layers to soften hard targets. Considering probability distributions at multiple network levels enhances small model generalization, leading to improved performance on unseen data.

\begin{figure*}[!t]
\centering
\includegraphics[width=\linewidth]{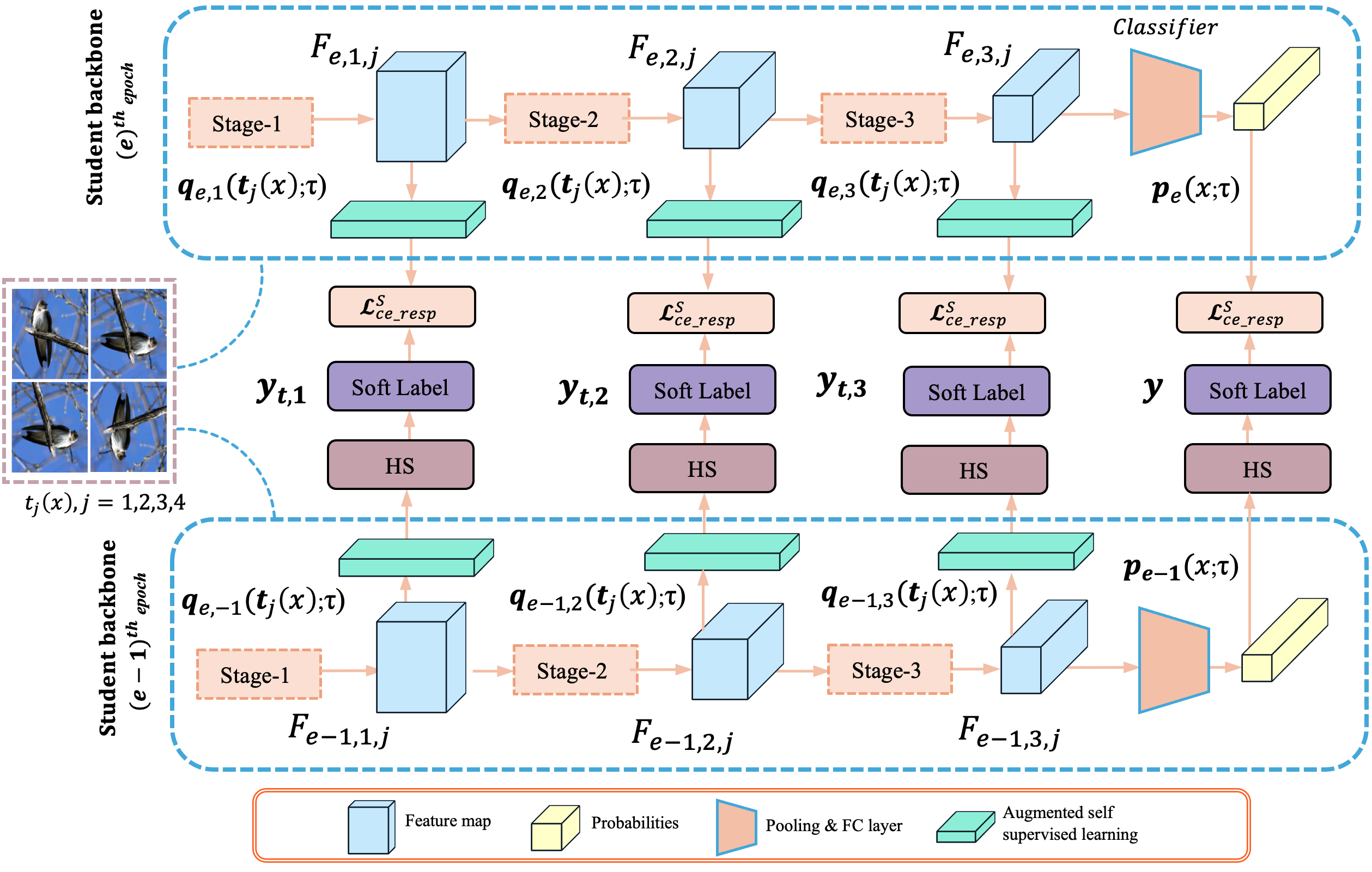}
\caption{Temporal self-knowledge distillation: the student backbone at epoch $e$ mimics the predictions of its own $(e{-}1)$-th epoch version.}
\label{fig_2}
\end{figure*}

\section{Methods}
\subsection{Overview}

The proposed architecture is shown in Figure~\ref{fig_1}. We explain the hierarchical consecutive learning of the student network using the predictive class distributions and SAD. 

To compute the predictive distribution, let $x \in X$ and $y \in \{1, \ldots, N\}$ be a training sample and target space with $N$ classes respectively. We train a CNN, denoted as $f(\cdot)$, to map input samples $x$ to predictive class probabilities $p \in \mathbb{R}^N$. The CNN consists of a feature extractor $\Phi(\cdot)$ and a linear classifier $W$, which encodes the feature embedding vector, and the classifier $W$ transforms it into a class logit probability distribution: $f(x) = W^\top \Phi(x) \in \mathbb{R}^N$. To produce soft probabilities with reduced class confidence, we define the predictive class probability with softmax temperature as $p(x; \tau) = \sigma\left(\frac{g(z; w)}{\tau}\right) \in \mathbb{R}^N$~\cite{hinton2015distilling}, where $\tau$ is the temperature hyper-parameter, and $w \in \mathbb{R}^{d \times N}$.

For the self-supervised task, we utilize the joint distribution of the original labels and the self-supervised augmented labels (SAL). During training with the dataset, there are $N$ labels that undergo $M$ transformations. Hence, the unified label space becomes $N \times M$, where $\times$ denotes element-wise multiplication. For instance, we train the CIFAR-100 (100 labels) with rotation transformation as self-supervision, which constructs $M = 4$ rotated transformations of images (0°, 90°, 180°, 270°), then it learns the joint distribution on all possible combinations, i.e., 400 labels.

Let $x' = \{t_j\}_{j=1}^M$ denote the self-supervised augmented training sample using transformation $t$. To define the SAD with softmax temperature, we employ the following approach:

\[q(\tilde{x}; \tau) = \sigma\left(\frac{g(z; w)}{\tau}\right) \in \mathbb{R}^{N \times M},\] where $\tau$ is the temperature hyper-parameter, and $w \in \mathbb{R}^{d \times N^*M}$.

The modern CNN uses stage-wise building blocks to learn features at various levels of abstraction. To seek the stage-wise representational supervision from the network, we choose to append an auxiliary branch after each stage, thus resulting in $L$ branches $\{c_l(\cdot)\}_{l=1}^L$, where $c_l(\cdot)$ denotes the auxiliary branch after the $l$th stage, inspired by~\cite{chen2022knowledge}. However, appending linear auxiliary classifiers without extra feature extraction modules may not effectively explore meaningful information, as empirically verified by~\cite{he2016identity}. Therefore, at each auxiliary branch contains feature extraction module, a global average pooling layer, and a linear classifier with a desirable dimension for the used auxiliary task, e.g., $N \times M$ for our self-supervision augmented task. We use $p_e(x; \tau)$ to denote the original class distribution at epoch $e$, $q_{e,l}(x; \tau)$ to denote the SAD of the $l$th stage at epoch $e$, and $q_{e-1,l}(t_j(x); \tau)$ to denote the SAD at the previous epoch, and $f_S(\cdot)$ to denote the student backbone.

\subsection{Consecutive Self Knowledge Distillation}

In our approach, we leverage predictions from past epochs to progressively soften the hard targets for subsequent epochs. Specifically, for the predictive class distribution obtained from the deepest classifier, we compute the soft targets using the following equation:

\begin{equation}
    y^s = (1 - \alpha)y^* + \alpha p_S^{e-1}(x;\tau),
\end{equation} where $p_S^{e-1}(x;\tau)$ represents the predictive class probability distribution from the previous epoch, which helps soften the target $y^s$. Here, $y^s$ represents the one-hot encoded label for the $N$ classes.

\begin{figure}
    \centering
    \includegraphics[width=0.9\linewidth]{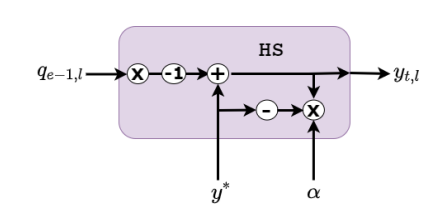}
    \caption{Architecture of the hybrid self-supervised (HS) fusion module for auxiliary classifier outputs. The previous epoch's distribution $q_{e-1,l}$ is adjusted using the target $y^*$ and weighted by factor $\alpha$ to obtain the current prediction $y_{t,l}$.}
    \label{fig3}
\end{figure}

For the self-supervised augmented distributions obtained from the $l$-th stage shallow classifiers, we have the following equation for converting hard labels to soft labels:

\begin{equation}
    y_{S,t,l} = \sum_{j=1}^{M} \left((1 - \alpha) y_{\ast,t} + \alpha q_{S}^{e-1,l}(t_j(x); \tau)\right),
\end{equation} where $q_{S}^{e-1,l}(t_j(x); \tau)$ represents the self-supervised augmented distribution of the $l$-th stage auxiliary classifier from the previous epoch. Here, $y_{\ast,t}$ denotes the transformed one-hot encoder for the $N$ classes, as shown in Figure~\ref{fig3}. With $M$ transformations, the indicator function of $y_{\ast,t}$ can be described as follows:

\begin{equation}
    y_{\ast,t} = \frac{1}{K} {1}_{\{\tilde{x}_i \in K\}} = \begin{cases}
    1, & \text{when } \tilde{x}_i \in K \\
    0, & \text{otherwise}
    \end{cases},
\end{equation} where $\tilde{x}$ represents the transformed sample, and $K = N \times M$, representing the label space for the transformed sample.

\subsection{Training the Student Network}

In our training procedure, we employ an end-to-end approach to enable the student model to distill its knowledge from both intermediate and output layers, enhancing its generalization capacity. Figure~\ref{fig_2} provides a detailed diagrammatic representation of our method.

Firstly, we compute two losses based on the output of intermediate layers and the final layer. For the predictive class distribution obtained from the deepest classifier, denoted as $p_S^e(x; \tau)$, we calculate the soft targets to obtain more informative labels $y_S$, using the cross-entropy loss function:
\begin{equation}
    \mathcal{L}_{S \text{ce resp}} = \mathcal{L}_{\text{ce}}\left(p_S^e(x; \tau), y_S\right),
\end{equation} where $\mathcal{L}_{S \text{ce resp}}$ represents the responsive cross-entropy loss for the student model. Here, $\tau = 1$, and $y_S$ denotes the soft targets derived from previous predictions $p_{S}^{e-1}(x; \tau)$.

Secondly, we feed the transformed samples $x^{\sim} = \{t_j\}_{j=1}^{M}$ into the student backbone $f_S(\cdot)$ to obtain transformed feature maps at each stage. These feature maps are then fed into $L$ auxiliary classifiers $c_{S,l}(\cdot)$ to compute the self-supervised augmented distribution $q_{S}^{e-1,l}(t_j(x); \tau)$ at the $(e-1)$th epoch, softening the hard targets for the $e$th epoch. The hierarchical loss with self-supervised softened labels across $M$ transformations is calculated as:

\begin{equation}
    \mathcal{L}_{S, \text{ce heir SAD}} = \frac{1}{M} \sum_{j=1}^{M} \sum_{l=1}^{L} \mathcal{L}_{\text{ce}} \left( q_{S}^{e,l}(t_j(x); \tau), y_{S} \right),
\end{equation} where $\mathcal{L}_{S, \text{ce heir SAD}}$ denotes the hierarchical cross-entropy loss between the SAD and the soft label. The label $y_{S,t,l}^{(2)}$ represents the joint softened distribution of the original supervised labels ($N$) and the self-supervised augmented labels ($M$).

Finally, we combine the two losses to obtain the Label-Supervised (LS) Loss, which mimics the soft label computed by smoothing the hard targets:

\begin{equation}
    \mathcal{L}^S_{LS} = \mathcal{L}^S_{\text{ce-resp}} + \mathcal{L}^S_{\text{ce-heir-SAD}}.
\end{equation}

\subsection{Feature and Classifier-based Distillation}

We incorporate two additional losses to leverage knowledge from later stages of the network. Firstly, to distill knowledge from later classifiers to earlier ones, we introduce a KL divergence loss. This loss encourages the shallow classifiers to mimic the self-supervised augmented distribution generated by the deeper classifier:

\begin{equation}
    \mathcal{L}^S_{\text{div-SAD}} = \frac{1}{M} \sum_{j=1}^{M} \sum_{l=1}^{L} \tau_{2DKL}\left(q_S^l(t_j(x);\tau) \Big\| q_S^O(t_j(x);\tau)\right)
\end{equation}

Here, $q_S^O(t_j(x);\tau)$ represents the self-supervised augmented distribution generated by the deeper classifier.

Secondly, we exploit the richer representational information contained in feature maps for knowledge distillation. Specifically, we compute the $\mathcal{L}_2$ loss function between the self-supervised augmented feature maps from the bottleneck stages and the feature maps from the last layer:

\begin{equation}
    \mathcal{L}_S^{\text{feat}} = \left\| F_t^l - F_t^o \right\|_2^2
\end{equation}

Here, $F_t^l$ represents the feature map of the average pooling layer of the $l$-th bottleneck stage, $F_t^o$ represents the feature map of the average pooling layer of the last layer, and $t$ denotes the transformed data sample fed to the $f_S(\cdot)$ backbone.

The hierarchical structure of our LSSKD framework aligns well with the staged information processing observed in biological perception systems and robotic agents. Furthermore, its self-supervised transformation mechanism can be extended to heterogeneous sensor modalities (e.g., radar, vision, and audio) through the design of modality-specific augmentation strategies, making LSSKD a promising candidate for multimodal sensor fusion tasks.

We combine these two losses and refer to them as the Itself-Supervised (IS) Loss, as they are both supervised using the knowledge of the network itself:

\begin{equation}
    \mathcal{L}_S^{\text{IS}} = \mathcal{L}_S^{\text{div}} + \mathcal{L}_S^{\text{feat}}
\end{equation}

Finally, we sum up the Label-Supervised Loss and the Itself-Supervised Loss to obtain the total loss, with hyperparameters $\beta$ and $\gamma$ assigning weights to each loss:

\begin{equation*}
    \mathcal{L}_T = \mathcal{L}_S^{\text{LS}} + \mathcal{L}_S^{\text{IS}} = (1 - \beta) \cdot \mathcal{L}_S^{\text{ce resp}} + \mathcal{L}_S^{\text{ce heir}} + \beta \cdot \mathcal{L}_S^{\text{div}} + \gamma \cdot \mathcal{L}_S^{\text{feat}},
\end{equation*}
the hyperparameters $\beta$ and $\gamma$ assign weights to each loss based on their optimization contributions. This process refines feature maps and auxiliary classifiers using both soft and hard labels, while ensuring earlier stages mimic later ones. During inference, all auxiliary branches can be removed, eliminating any extra computational burden.

\section{Experimental Results}

\subsection{Experimental Settings}

All experiments were conducted on an NVIDIA RTX 4090 GPU. We evaluated the effectiveness and generalizability of our proposed LSSKD framework on three benchmark datasets: CIFAR-100~\cite{krizhevsky2009learning}, ImageNet~\cite{krizhevsky2012imagenet}, and Tiny-ImageNet~\cite{le2015tiny}. A diverse set of teacher–student architectures was used, including PreAct ResNet-18~\cite{he2016identity}, ResNet~\cite{he2016deep}, WRN~\cite{zagoruyko2016wide}, ResNeXt~\cite{xie2017aggregated}, VGG~\cite{simonyan2014very}, MobileNetV2~\cite{sandler2018mobilenetv2}, and ShuffleNet~\cite{ma2018shufflenet, zhang2018shufflenet}, as summarized in Table~\ref{tab:my_label1}. On ImageNet, we employed ResNet-18 for both teacher and student to validate scalability, with results shown in Table~\ref{tab:my_label2}.

We applied standard data augmentation strategies following~\cite{he2016deep}, including 4-pixel padding, random cropping to $32\times32$, horizontal flipping, and normalization. All models were trained for 240 epochs using stochastic gradient descent (SGD) with a momentum of 0.9. The initial learning rate was set to 0.05 and decayed by a factor of 10 at the 150th and 210th epochs. The batch size was 64, and the weight decay was set to $5 \times 10^{-5}$. 
LSSKD introduces three hyperparameters: $\alpha$, $\beta$, and $\gamma$, which respectively control the contributions of softened labels, auxiliary supervision, and feature-level consistency. These values were tuned using a 10\% held-out validation split from the training data. The final values used in all experiments were $\alpha = 0.8$, $\beta = 0.1$, and $\gamma = 0.1$.

\subsection{Overall Performance}
Our proposed LSSKD achieves state-of-the-art performance on CIFAR-100 across various student-teacher architectures without requiring any pre-trained teacher model. As shown in Table~\ref{tab:my_label1}, LSSKD consistently outperforms prior KD methods, including SSKD~\cite{xu2020knowledge}, FitNet~\cite{adriana2015fitnets}, and CRD~\cite{tian2019contrastive}.
Specifically, LSSKD achieves an average improvement of 4.54\% over SSKD on CIFAR-100. On ImageNet, it yields a top-1 gain of 0.33\% using ResNet-18 as both teacher and student (see Table~\ref{tab:my_label2}).
To ensure consistent and fair evaluation, we apply the same training pipeline and data augmentations across all methods, including random cropping, horizontal flipping, and SGD training with identical learning rate schedules.
These results verify the superiority of LSSKD in achieving high accuracy while maintaining training and inference efficiency.


\begin{figure}[ht]
    \centering
    \includegraphics[width=0.9\linewidth]{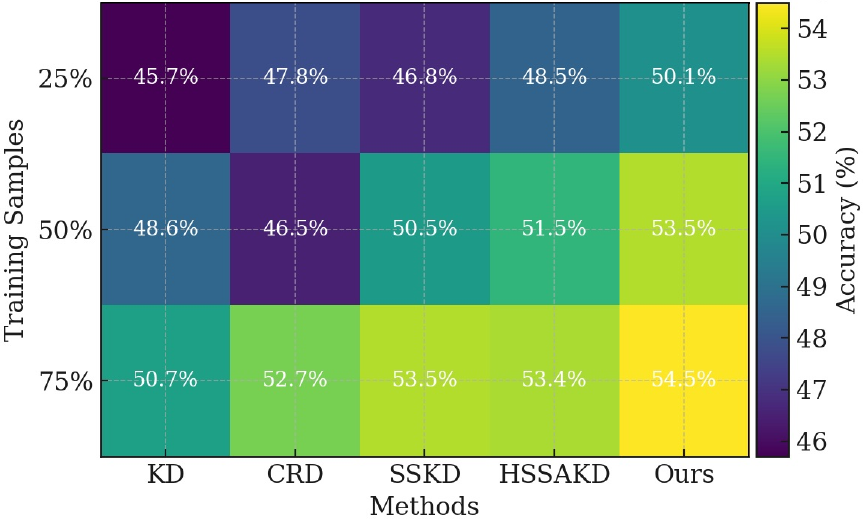}
    \caption{Top-1 accuracy (\%) comparison on Tiny-ImageNet under the few-shot scenario using different proportions of training data. ResNet56–ResNet20 is used as the teacher–student pair.}
    \label{fig:my_label41}
\end{figure}

\begin{figure}[ht]
    \centering
    \includegraphics[width=0.9\linewidth]{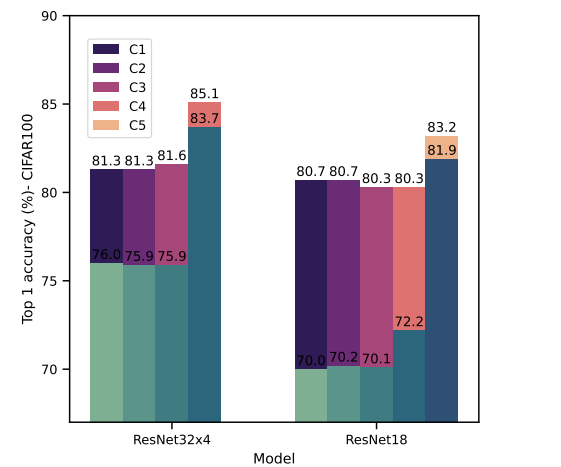}
    \caption{Stagewise comparison with HASKD~\cite{yang2021hierarchical}. Red bars denote the accuracy of HASKD, while blue bars represent our method. C4 (R32$\times$4) and C5 (R18) are final classifiers; others are auxiliary classifiers.}
    \label{fig:my_label12315}
\end{figure}

\begin{table*}[]
    \centering
    \label{tab:my_label2}
    \renewcommand{\arraystretch}{0.9} 
    \setlength{\tabcolsep}{2.2pt} 
    \caption{
    Performance of LSSKD compared with baseline methods on CIFAR-100. Teacher network: ResNet34; Student network: ResNet18. In other comparisons, we used the same ResNet34-ResNet18 pair to ensure consistency. All experiments were conducted with a batch size of 64 and a learning rate schedule decreasing by a factor of 10 at the 150th and 210th epochs.}
    \begin{tabular}{ccc|cccccccccc}
    \toprule
         \textbf{Accuracy} & \textbf{Teacher} & \textbf{Student} & KD~\cite{hinton2015distilling} & AT~\cite{zagoruyko2016paying} & CC~\cite{peng2019correlation} & SP~\cite{tung2019similarity} & RKD~\cite{park2019relational} & CRD~\cite{tian2019contrastive} & SSKD~\cite{xu2020knowledge} & HSSAKD~\cite{9827964} & MOKD~\cite{song2023multi} & Ours \\ 
         \midrule
         Top-1 & 73.31 & \cellcolor{gray!20}69.75 & 70.66 & 70.70 & 69.96 & 70.62 & 71.34 & 71.38 & 71.62 & 72.16 & \underline{72.3} & \textbf{72.45} \\
         Top-5 & 91.42 & \cellcolor{gray!20}89.07 & 89.88 & 90.0 & 89.17 & 89.80 & 90.37 & 90.49 & 90.67 & 90.85 & \underline{90.9} & \textbf{91.15} \\
         \bottomrule
    \end{tabular}
\end{table*}

\begin{table*}[]
    \centering
     \caption{Top-1 accuracy (\%) comparison of our self-distillation method with distillation methods across various teacher-student networks on CIFAR-100. The bold number represents the best accuracy. Data from ~\cite{9827964} has been added to provide a comprehensive comparison.}
    \label{tab:my_label1}
    \setlength{\tabcolsep}{2.5pt} 
    \begin{tabular}{rc|cccccc}
    \toprule
         \textbf{Distillation Mechanism}&\textbf{Teacher}& WRN-40-2& WRN-40-2& Resnet56& ResNet32x4& WRN-40-2&ResNet32x4  \\ 
         &\textbf{Student}& WRN-16-2 &WRN-40-1& Resnet20& ResNet8x4& ShuffleNetV1& ShuffleNetV2 \\ 
         \midrule
         \rowcolor{gray!20}&Baseline&73.57 &71.95& 69.62& 72.95& 71.74& 72.96 \\
         KD~\cite{hinton2015distilling} &&75.23& 73.90& 70.91& 73.54& 75.83& 75.43  \\
         FitNet~\cite{adriana2015fitnets}&& 75.30 &74.30& 71.21& 75.37& 76.27& 76.91  \\
AT~\cite{zagoruyko2016paying}&& 75.64& 74.32& 71.35& 75.06& 76.51& 76.32 \\ 
AB~\cite{heo2019knowledge} &&71.26& 74.55& 71.56& 74.31& 76.43& 76.40 \\ 
VID~\cite{ahn2019variational}&& 75.31& 74.23& 71.35& 75.07& 76.24 &75.98 \\ 
RKD~\cite{park2019relational}&& 75.33& 73.90& 71.67& 74.17& 75.74&75.42  \\
SP~\cite{tung2019similarity}&& 74.35& 72.91& 71.45& 75.44& 76.40& 76.43 \\  
CC~\cite{peng2019correlation} &&75.30& 74.46& 71.44& 74.40& 75.63& 75.74 \\  
CRD~\cite{tian2019contrastive} &&75.81& 74.76& \underline{71.83}& 75.77& 76.37& 76.51 \\ 
SSKD~\cite{xu2020knowledge}&& \underline{76.16}& \underline{75.84}& 70.80& 75.83& 76.71& 77.64 \\
SemCKD~\cite{Chen2020CrossLayerDW}&&75.02&73.54&71.54&\underline{75.89}&\underline{77.39}&\underline{78.26}\\
\textbf{Proposed Method}&&\textbf{76.89}& \textbf{78.35}& \textbf{72.67}&\textbf{76.27}&\textbf{77.45} &\textbf{79.43} \\
\bottomrule
    \end{tabular}
\end{table*}

\subsection{Generalization under Few-shot Settings}

To assess the generalization capability of LSSKD, we evaluate its performance under few-shot learning scenarios using CIFAR-100 and Tiny-ImageNet. In each case, we retain 25\%, 50\%, and 75\% of the training samples, while keeping the full test set intact. A stratified sampling strategy ensures balanced class representation across different fractions.
As shown in Figure~\ref{fig:my_label41}, LSSKD significantly outperforms conventional KD methods such as KD~\cite{hinton2015distilling}, CRD~\cite{tian2019contrastive}, and SSKD~\cite{xu2020knowledge} across all data scarcity levels. These results demonstrate the robustness and adaptability of our framework, even in low-data regimes—a critical property for real-world deployment where labeled data is limited.

\subsection{Comparison with State-of-the-Art Self-Distillation Methods}

We conduct a comprehensive comparison of LSSKD against recent self-distillation methods, including LS~\cite{szegedy2016rethinking}, CSKD~\cite{yun2020regularizing}, TFKD~\cite{yuan2020revisiting}, and PS-KD~\cite{kim2021self}. All evaluations are performed on CIFAR-100 using a consistent ResNet34–ResNet18 architecture for fairness. The results are presented in Table~\ref{tab:my_label2311}. 
In addition to vanilla self-distillation, we explore the impact of data augmentation strategies—specifically, Cutout and CutMix (COC)~\cite{devries2017improved, yun2019cutmix}—on performance. LSSKD consistently outperforms all baselines, achieving an average gain of 4.56\% over PS-KD and an additional boost of 2.58\% when enhanced with COC. Notably, on lightweight models such as MobileNetV2 and ShuffleNetV2, our method yields improvements of up to 5.6\% in top-1 accuracy, demonstrating superior regularization and scalability.

\section{Ablation Studies and Analysis}

\subsection{Impact of Auxiliary Classifiers}
We evaluate the role of intermediate auxiliary classifiers in enhancing the representational capacity of the student network. Inspired by HASKD~\cite{yang2021hierarchical}, LSSKD attaches stage-wise classifiers across bottleneck layers, with supervision derived from the deepest classifier’s soft labels. As illustrated in Figure~\ref{fig:my_label12315}, our method significantly improves classification accuracy at each stage compared to HASKD.
The observed gains validate two key assumptions: (1) hierarchical supervision via softened labels is more informative than conventional hard targets, and (2) the deepest classifier provides stronger generalization, making it ideal for supervising shallower modules. Empirical results on ResNet-18 and ResNet-32x4 confirm notable performance improvements across all auxiliary classifiers.

\begin{table}[]
    \centering
    \caption{Top-1 accuracy (\%) comparison of SOTA self-distillation
methods across various student networks on CIFAR-100.}
    \label{tab:my_label2311}
    \setlength{\tabcolsep}{1.5pt} 
    \begin{tabular}{cccc}
         \toprule \textbf{Method}& ResNet18& ResNet101& MobileNetV2 \\ \midrule
         \rowcolor{gray!20} Baseline& 75.87& 79.25& 68.38 \\
LS~\cite{szegedy2016rethinking}& 79.06& 80.16& $-$ \\ 
CSKD~\cite{yun2020regularizing}& 78.7& 79.53&$ -$ \\ 
TFKD~\cite{yuan2020revisiting}& 77.36& 77.36& 70.88 \\ 
$TFKD_{self}$& 77.11& 77.11& \underline{70.96} \\ 
PS-KD~\cite{kim2021self}& \underline{79.18}& \underline{80.57}& 69.89 \\ 
\textbf{Proposed MSAKD}&\textbf{83.16}&\textbf{82.24} &\textbf{73.95} \\ 
\bottomrule
    \end{tabular}
\end{table}

\begin{table}[]
    \centering
     \caption{Baseline comparison. The number that is written in subscript form shows the increase of accuracy over the baseline method i.e. 72.67(+3.05) shows an increment of the 3.05\% over the baseline accuracy of 69.62\%.}
    \label{tab:my_label00}
    \begin{tabular}{ccc}
    \toprule
        \textbf{Method} & \textbf{Baseline}& \textbf{Proposed MSAKD}  \\
        \midrule
         ResNet20& 69.62& 72.67(+3.05) \\ 
       ResNet56 &71.83& 78.79(+3.96) \\ \midrule
       WRN-16-2& 76.77& 83.16(+6.39) \\ 
       WRN-40-2&76.89& 85.68(+8.79) \\ \midrule
       ResNet18& 73.57& 76.89(+3.32) \\ 
       ResNet50 &75.81& 82.63(+6.22) \\ \bottomrule
    \end{tabular}
\end{table}


\subsection{Effectiveness of Hierarchical Softened Labels}
Unlike prior methods such as PS-KD~\cite{kim2021self}, which rely solely on final-layer predictions to soften hard targets, our LSSKD introduces a hierarchical softening strategy using intermediate classifiers. Each layer’s probabilistic output contributes to refining the supervision signals, forming a multi-stage softened label.
Table~\ref{tab:my_label00} demonstrates that this hierarchical strategy consistently improves student performance across multiple architectures. For instance, our method yields accuracy gains of up to 8.79\% on WRN-40-2 and 6.39\% on WRN-16-2, showing that softened hierarchical supervision enhances optimization and generalization compared to baseline hard-label training.

\subsection{Robustness across Network Architectures}
To assess the general applicability of LSSKD, we evaluated its performance across various baseline architectures, including ResNet, WRN, and ShuffleNet families. As shown in Table~\ref{tab:my_label00}, our method consistently improves accuracy regardless of network depth or complexity.
For example, LSSKD achieves a gain of +3.05\% on ResNet20, +3.96\% on ResNet56, and up to +8.79\% on WRN-40-2. These results confirm the robustness and versatility of LSSKD in enhancing diverse model backbones, making it suitable for both lightweight and deep neural networks.

\section{Limitations}
Despite the strong performance of LSSKD, several aspects warrant further exploration:
\textbf{(1) Scalability.} The effectiveness of LSSKD on extremely large-scale datasets across broader domains remains underexplored.
\textbf{(2) Task Generalization.} Future work will investigate its extension to tasks beyond image classification, such as object detection, machine translation, and model compression.
\textbf{(3) Model Simplification.} Further efforts can focus on reducing model complexity by pruning deeper layers, enhancing the method’s applicability in resource-constrained environments.

\section{Conclusion}
We present the LSSKD framework, which improves model performance through knowledge distillation from both intermediate and final layers. Our method achieves state-of-the-art results on CIFAR-100 and ImageNet without relying on large pre-trained teacher networks. Notably, all auxiliary classifiers are removed during inference, ensuring no additional computational cost. The efficiency and compactness of LSSKD make it especially advantageous for deploying small language models on cost-effective devices, enabling effective training with fewer parameters and minimal resource demands. This supports robust and accurate models in privacy-sensitive and low-latency scenarios. Moreover, by distilling large models into compact versions, LSSKD reduces the carbon footprint and broadens access to high-performance AI beyond major technology companies. In future work, we plan to explore its applicability to tasks such as object detection, machine translation, and model compression.

\bibliographystyle{IEEEtran}
\bibliography{IEEE/Ref}

\begin{thebibliography}{10}
\providecommand{\url}[1]{#1}
\csname url@samestyle\endcsname
\providecommand{\newblock}{\relax}
\providecommand{\bibinfo}[2]{#2}
\providecommand{\BIBentrySTDinterwordspacing}{\spaceskip=0pt\relax}
\providecommand{\BIBentryALTinterwordstretchfactor}{4}
\providecommand{\BIBentryALTinterwordspacing}{\spaceskip=\fontdimen2\font plus
\BIBentryALTinterwordstretchfactor\fontdimen3\font minus \fontdimen4\font\relax}
\providecommand{\BIBforeignlanguage}[2]{{%
\expandafter\ifx\csname l@#1\endcsname\relax
\typeout{** WARNING: IEEEtran.bst: No hyphenation pattern has been}%
\typeout{** loaded for the language `#1'. Using the pattern for}%
\typeout{** the default language instead.}%
\else
\language=\csname l@#1\endcsname
\fi
#2}}
\providecommand{\BIBdecl}{\relax}
\BIBdecl

\bibitem{mikolajczyk2018data}
A.~Miko{\l}ajczyk and M.~Grochowski, ``Data augmentation for improving deep learning in image classification problem,'' in \emph{2018 international interdisciplinary PhD workshop (IIPhDW)}.\hskip 1em plus 0.5em minus 0.4em\relax IEEE, 2018, pp. 117--122.

\bibitem{yu2024capan}
Z.~Yu and P.~Wang, ``Capan: Class-aware prototypical adversarial networks for unsupervised domain adaptation,'' in \emph{2024 IEEE International Conference on Multimedia and Expo (ICME)}.\hskip 1em plus 0.5em minus 0.4em\relax IEEE, 2024, pp. 1--6.

\bibitem{wang2024multi}
P.~Wang, Y.~Yang, and Z.~Yu, ``Multi-batch nuclear-norm adversarial network for unsupervised domain adaptation,'' in \emph{2024 IEEE International Conference on Multimedia and Expo (ICME)}.\hskip 1em plus 0.5em minus 0.4em\relax IEEE, 2024, pp. 1--6.

\bibitem{zhao2019object}
Z.-Q. Zhao, P.~Zheng, S.-t. Xu, and X.~Wu, ``Object detection with deep learning: A review,'' \emph{IEEE transactions on neural networks and learning systems}, vol.~30, no.~11, pp. 3212--3232, 2019.

\bibitem{huang2022evaluation}
Y.~Huang, J.~Wu, X.~Xu, and S.~Ding, ``Evaluation-oriented knowledge distillation for deep face recognition,'' in \emph{Proceedings of the IEEE/CVF Conference on Computer Vision and Pattern Recognition}, 2022, pp. 18\,740--18\,749.

\bibitem{zhang2019fast}
F.~Zhang, X.~Zhu, and M.~Ye, ``Fast human pose estimation,'' in \emph{Proceedings of the IEEE/CVF conference on computer vision and pattern recognition}, 2019, pp. 3517--3526.

\bibitem{liu2021semantics}
Y.~Liu, K.~Wang, G.~Li, and L.~Lin, ``Semantics-aware adaptive knowledge distillation for sensor-to-vision action recognition,'' \emph{IEEE Transactions on Image Processing}, vol.~30, pp. 5573--5588, 2021.

\bibitem{garcia2018survey}
A.~Garcia-Garcia, S.~Orts-Escolano, S.~Oprea, V.~Villena-Martinez, P.~Martinez-Gonzalez, and J.~Garcia-Rodriguez, ``A survey on deep learning techniques for image and video semantic segmentation,'' \emph{Applied Soft Computing}, vol.~70, pp. 41--65, 2018.

\bibitem{luo2022pixel}
Y.~Luo, J.~Wang, X.~Yang, Z.~Yu, and Z.~Tan, ``Pixel representation augmented through cross-attention for high-resolution remote sensing imagery segmentation,'' \emph{Remote Sensing}, vol.~14, no.~21, p. 5415, 2022.

\bibitem{thompson2020computational}
N.~C. Thompson, K.~Greenewald, K.~Lee, and G.~F. Manso, ``The computational limits of deep learning,'' \emph{arXiv preprint arXiv:2007.05558}, 2020.

\bibitem{yu2021compute}
S.~Yu, H.~Jiang, S.~Huang, X.~Peng, and A.~Lu, ``Compute-in-memory chips for deep learning: Recent trends and prospects,'' \emph{IEEE circuits and systems magazine}, vol.~21, no.~3, pp. 31--56, 2021.

\bibitem{strubell2020energy}
E.~Strubell, A.~Ganesh, and A.~McCallum, ``Energy and policy considerations for modern deep learning research,'' in \emph{Proceedings of the AAAI conference on artificial intelligence}, vol.~34, no.~09, 2020, pp. 13\,693--13\,696.

\bibitem{ji2024edge}
J.~Ji, Z.~Shu, H.~Li, K.~X. Lai, M.~Lu, G.~Jiang, W.~Wang, Y.~Zheng, and X.~Jiang, ``Edge-computing based knowledge distillation and multi-task learning for partial discharge recognition,'' \emph{IEEE Transactions on Instrumentation and Measurement}, 2024.

\bibitem{yin2024dpal}
L.~Yin, L.~Wang, Z.~Cai, S.~Lu, R.~Wang, A.~AlSanad, S.~A. AlQahtani, X.~Chen, Z.~Yin, X.~Li \emph{et~al.}, ``Dpal-bert: A faster and lighter question answering model.'' \emph{CMES-Computer Modeling in Engineering \& Sciences}, vol. 141, no.~1, 2024.

\bibitem{han2015learning}
S.~Han, J.~Pool, J.~Tran, and W.~Dally, ``Learning both weights and connections for efficient neural network,'' \emph{Advances in neural information processing systems}, vol.~28, 2015.

\bibitem{wu2016quantized}
J.~Wu, C.~Leng, Y.~Wang, Q.~Hu, and J.~Cheng, ``Quantized convolutional neural networks for mobile devices,'' in \emph{Proceedings of the IEEE conference on computer vision and pattern recognition}, 2016, pp. 4820--4828.

\bibitem{buciluǎ2006model}
C.~Buciluǎ, R.~Caruana, and A.~Niculescu-Mizil, ``Model compression,'' in \emph{Proceedings of the 12th ACM SIGKDD international conference on Knowledge discovery and data mining}, 2006, pp. 535--541.

\bibitem{kim2021self}
K.~Kim, B.~Ji, D.~Yoon, and S.~Hwang, ``Self-knowledge distillation with progressive refinement of targets,'' in \emph{Proceedings of the IEEE/CVF International Conference on Computer Vision}, 2021, pp. 6567--6576.

\bibitem{nowlan2018simplifying}
S.~J. Nowlan and G.~E. Hinton, ``Simplifying neural networks by soft weight sharing,'' in \emph{The Mathematics of Generalization}.\hskip 1em plus 0.5em minus 0.4em\relax CRC Press, 2018, pp. 373--394.

\bibitem{srivastava2014dropout}
N.~Srivastava, G.~Hinton, A.~Krizhevsky, I.~Sutskever, and R.~Salakhutdinov, ``Dropout: a simple way to prevent neural networks from overfitting,'' \emph{The journal of machine learning research}, vol.~15, no.~1, pp. 1929--1958, 2014.

\bibitem{ioffe2015batch}
S.~Ioffe and C.~Szegedy, ``Batch normalization: Accelerating deep network training by reducing internal covariate shift,'' in \emph{International conference on machine learning}.\hskip 1em plus 0.5em minus 0.4em\relax pmlr, 2015, pp. 448--456.

\bibitem{devries2017improved}
T.~DeVries and G.~W. Taylor, ``Improved regularization of convolutional neural networks with cutout,'' \emph{arXiv preprint arXiv:1708.04552}, 2017.

\bibitem{hinton2015distilling}
G.~Hinton, O.~Vinyals, and J.~Dean, ``Distilling the knowledge in a neural network,'' \emph{arXiv preprint arXiv:1503.02531}, 2015.

\bibitem{yu2024improved}
Z.~Yu, ``Improved implicit diffusion model with knowledge distillation to estimate the spatial distribution density of carbon stock in remote sensing imagery,'' \emph{arXiv preprint arXiv:2411.17973}, 2024.

\bibitem{zagoruyko2016paying}
S.~Zagoruyko and N.~Komodakis, ``Paying more attention to attention: Improving the performance of convolutional neural networks via attention transfer,'' \emph{arXiv preprint arXiv:1612.03928}, 2016.

\bibitem{yang2021hierarchical}
C.~Yang, Z.~An, L.~Cai, and Y.~Xu, ``Hierarchical self-supervised augmented knowledge distillation,'' \emph{arXiv preprint arXiv:2107.13715}, 2021.

\bibitem{yang2022knowledge}
------, ``Knowledge distillation using hierarchical self-supervision augmented distribution,'' \emph{IEEE Transactions on Neural Networks and Learning Systems}, 2022.

\bibitem{sarfraz2021knowledge}
F.~Sarfraz, E.~Arani, and B.~Zonooz, ``Knowledge distillation beyond model compression,'' in \emph{2020 25th International Conference on Pattern Recognition (ICPR)}.\hskip 1em plus 0.5em minus 0.4em\relax IEEE, 2021, pp. 6136--6143.

\bibitem{doshi2024reffakd}
D.~Doshi and J.-E. Kim, ``Reffakd: Resource-efficient autoencoder-based knowledge distillation,'' \emph{arXiv preprint arXiv:2404.09886}, 2024.

\bibitem{yang2025constructing}
J.~Yang, Z.~Liu, G.~Wang, Q.~Zhang, S.~Xia, D.~Wu, and Y.~Liu, ``Constructing three-way decision with fuzzy granular-ball rough sets based on uncertainty invariance,'' \emph{IEEE Transactions on Fuzzy Systems}, 2025.

\bibitem{lopez2015unifying}
D.~Lopez-Paz, L.~Bottou, B.~Sch{\"o}lkopf, and V.~Vapnik, ``Unifying distillation and privileged information,'' \emph{arXiv preprint arXiv:1511.03643}, 2015.

\bibitem{chen2020online}
D.~Chen, J.-P. Mei, C.~Wang, Y.~Feng, and C.~Chen, ``Online knowledge distillation with diverse peers,'' in \emph{Proceedings of the AAAI conference on artificial intelligence}, vol.~34, no.~04, 2020, pp. 3430--3437.

\bibitem{son2021densely}
W.~Son, J.~Na, J.~Choi, and W.~Hwang, ``Densely guided knowledge distillation using multiple teacher assistants,'' in \emph{Proceedings of the IEEE/CVF International Conference on Computer Vision}, 2021, pp. 9395--9404.

\bibitem{li2024continual}
S.~Li, T.~Su, X.~Zhang, and Z.~Wang, ``Continual learning with knowledge distillation: A survey,'' \emph{Authorea Preprints}, 2024.

\bibitem{rao2023dynamic}
J.~Rao, L.~Ding, S.~Qi, M.~Fang, Y.~Liu, L.~Shen, and D.~Tao, ``Dynamic contrastive distillation for image-text retrieval,'' \emph{IEEE Transactions on Multimedia}, vol.~25, pp. 8383--8395, 2023.

\bibitem{gou2021knowledge}
J.~Gou, B.~Yu, S.~J. Maybank, and D.~Tao, ``Knowledge distillation: A survey,'' \emph{International Journal of Computer Vision}, vol. 129, no.~6, pp. 1789--1819, 2021.

\bibitem{hadi2023survey}
M.~U. Hadi, R.~Qureshi, A.~Shah, M.~Irfan, A.~Zafar, M.~B. Shaikh, N.~Akhtar, J.~Wu, S.~Mirjalili \emph{et~al.}, ``A survey on large language models: Applications, challenges, limitations, and practical usage,'' \emph{Authorea Preprints}, 2023.

\bibitem{xu2024survey}
X.~Xu, M.~Li, C.~Tao, T.~Shen, R.~Cheng, J.~Li, C.~Xu, D.~Tao, and T.~Zhou, ``A survey on knowledge distillation of large language models,'' \emph{arXiv preprint arXiv:2402.13116}, 2024.

\bibitem{achiam2023gpt}
J.~Achiam, S.~Adler, S.~Agarwal, L.~Ahmad, I.~Akkaya, F.~L. Aleman, D.~Almeida, J.~Altenschmidt, S.~Altman, S.~Anadkat \emph{et~al.}, ``Gpt-4 technical report,'' \emph{arXiv preprint arXiv:2303.08774}, 2023.

\bibitem{touvron2023llama}
H.~Touvron, L.~Martin, K.~Stone, P.~Albert, A.~Almahairi, Y.~Babaei, N.~Bashlykov, S.~Batra, P.~Bhargava, S.~Bhosale \emph{et~al.}, ``Llama 2: Open foundation and fine-tuned chat models,'' \emph{arXiv preprint arXiv:2307.09288}, 2023.

\bibitem{karamcheti2021mistral}
S.~Karamcheti, L.~Orr, J.~Bolton, T.~Zhang, K.~Goel, A.~Narayan, R.~Bommasani, D.~Narayanan, T.~Hashimoto, D.~Jurafsky \emph{et~al.}, ``Mistral--a journey towards reproducible language model training,'' 2021.

\bibitem{cho2019efficacy}
J.~H. Cho and B.~Hariharan, ``On the efficacy of knowledge distillation,'' in \emph{Proceedings of the IEEE/CVF international conference on computer vision}, 2019, pp. 4794--4802.

\bibitem{kim2019curiosity}
Y.~Kim, W.~Nam, H.~Kim, J.-H. Kim, and G.~Kim, ``Curiosity-bottleneck: Exploration by distilling task-specific novelty,'' in \emph{International conference on machine learning}.\hskip 1em plus 0.5em minus 0.4em\relax PMLR, 2019, pp. 3379--3388.

\bibitem{hao2022learning}
Z.~Hao, J.~Guo, D.~Jia, K.~Han, Y.~Tang, C.~Zhang, H.~Hu, and Y.~Wang, ``Learning efficient vision transformers via fine-grained manifold distillation,'' \emph{Advances in Neural Information Processing Systems}, vol.~35, pp. 9164--9175, 2022.

\bibitem{chen2020robust}
T.~Chen, Z.~Zhang, S.~Liu, S.~Chang, and Z.~Wang, ``Robust overfitting may be mitigated by properly learned smoothening,'' in \emph{International Conference on Learning Representations}, 2020.

\bibitem{ahn2019variational}
S.~Ahn, S.~X. Hu, A.~Damianou, N.~D. Lawrence, and Z.~Dai, ``Variational information distillation for knowledge transfer,'' in \emph{Proceedings of the IEEE/CVF Conference on Computer Vision and Pattern Recognition}, 2019, pp. 9163--9171.

\bibitem{park2019relational}
W.~Park, D.~Kim, Y.~Lu, and M.~Cho, ``Relational knowledge distillation,'' in \emph{Proceedings of the IEEE/CVF Conference on Computer Vision and Pattern Recognition}, 2019, pp. 3967--3976.

\bibitem{tung2019similarity}
F.~Tung and G.~Mori, ``Similarity-preserving knowledge distillation,'' in \emph{Proceedings of the IEEE/CVF international conference on computer vision}, 2019, pp. 1365--1374.

\bibitem{adriana2015fitnets}
R.~Adriana, B.~Nicolas, K.~S. Ebrahimi, C.~Antoine, G.~Carlo, and B.~Yoshua, ``Fitnets: Hints for thin deep nets,'' \emph{Proc. ICLR}, vol.~2, no.~3, p.~1, 2015.

\bibitem{yim2017gift}
J.~Yim, D.~Joo, J.~Bae, and J.~Kim, ``A gift from knowledge distillation: Fast optimization, network minimization and transfer learning,'' in \emph{Proceedings of the IEEE conference on computer vision and pattern recognition}, 2017, pp. 4133--4141.

\bibitem{heo2019knowledge}
B.~Heo, M.~Lee, S.~Yun, and J.~Y. Choi, ``Knowledge transfer via distillation of activation boundaries formed by hidden neurons,'' in \emph{Proceedings of the AAAI Conference on Artificial Intelligence}, vol.~33, no.~01, 2019, pp. 3779--3787.

\bibitem{lee2015deeply}
C.-Y. Lee, S.~Xie, P.~Gallagher, Z.~Zhang, and Z.~Tu, ``Deeply-supervised nets,'' in \emph{Artificial intelligence and statistics}.\hskip 1em plus 0.5em minus 0.4em\relax Pmlr, 2015, pp. 562--570.

\bibitem{chen2022knowledge}
D.~Chen, J.-P. Mei, H.~Zhang, C.~Wang, Y.~Feng, and C.~Chen, ``Knowledge distillation with the reused teacher classifier,'' in \emph{Proceedings of the IEEE/CVF conference on computer vision and pattern recognition}, 2022, pp. 11\,933--11\,942.

\bibitem{luo2023knowledge}
S.~Luo, D.~Chen, and C.~Wang, ``Knowledge distillation with deep supervision,'' in \emph{2023 International Joint Conference on Neural Networks (IJCNN)}.\hskip 1em plus 0.5em minus 0.4em\relax IEEE, 2023, pp. 1--8.

\bibitem{tian2019contrastive}
Y.~Tian, D.~Krishnan, and P.~Isola, ``Contrastive representation distillation,'' \emph{arXiv preprint arXiv:1910.10699}, 2019.

\bibitem{liu2021sskd}
W.~Liu, S.~Nie, J.~Yin, R.~Wang, D.~Gao, and L.~Jin, ``Sskd: Self-supervised knowledge distillation for cross domain adaptive person re-identification,'' in \emph{2021 7th IEEE International Conference on Network Intelligence and Digital Content (IC-NIDC)}.\hskip 1em plus 0.5em minus 0.4em\relax IEEE, 2021, pp. 81--85.

\bibitem{chen2020simple}
T.~Chen, S.~Kornblith, M.~Norouzi, and G.~Hinton, ``A simple framework for contrastive learning of visual representations,'' in \emph{International conference on machine learning}.\hskip 1em plus 0.5em minus 0.4em\relax PMLR, 2020, pp. 1597--1607.

\bibitem{yuan2020revisiting}
L.~Yuan, F.~E. Tay, G.~Li, T.~Wang, and J.~Feng, ``Revisiting knowledge distillation via label smoothing regularization,'' in \emph{Proceedings of the IEEE/CVF Conference on Computer Vision and Pattern Recognition}, 2020, pp. 3903--3911.

\bibitem{yun2020regularizing}
S.~Yun, J.~Park, K.~Lee, and J.~Shin, ``Regularizing class-wise predictions via self-knowledge distillation,'' in \emph{Proceedings of the IEEE/CVF conference on computer vision and pattern recognition}, 2020, pp. 13\,876--13\,885.

\bibitem{niu2022scale}
J.-Y. Niu, Z.-H. Xie, Y.~Li, S.-J. Cheng, and J.-W. Fan, ``Scale fusion light cnn for hyperspectral face recognition with knowledge distillation and attention mechanism,'' \emph{Applied Intelligence}, vol.~52, no.~6, pp. 6181--6195, 2022.

\bibitem{xu2020knowledge}
G.~Xu, Z.~Liu, X.~Li, and C.~C. Loy, ``Knowledge distillation meets self-supervision,'' in \emph{European conference on computer vision}.\hskip 1em plus 0.5em minus 0.4em\relax Springer, 2020, pp. 588--604.

\bibitem{you2017learning}
S.~You, C.~Xu, C.~Xu, and D.~Tao, ``Learning from multiple teacher networks,'' in \emph{Proceedings of the 23rd ACM SIGKDD international conference on knowledge discovery and data mining}, 2017, pp. 1285--1294.

\bibitem{he2016identity}
K.~He, X.~Zhang, S.~Ren, and J.~Sun, ``Identity mappings in deep residual networks,'' in \emph{Computer Vision--ECCV 2016: 14th European Conference, Amsterdam, The Netherlands, October 11--14, 2016, Proceedings, Part IV 14}.\hskip 1em plus 0.5em minus 0.4em\relax Springer, 2016, pp. 630--645.

\bibitem{krizhevsky2009learning}
A.~Krizhevsky, G.~Hinton \emph{et~al.}, ``Learning multiple layers of features from tiny images,'' 2009.

\bibitem{krizhevsky2012imagenet}
A.~Krizhevsky, I.~Sutskever, and G.~E. Hinton, ``Imagenet classification with deep convolutional neural networks,'' \emph{Advances in neural information processing systems}, vol.~25, 2012.

\bibitem{le2015tiny}
Y.~Le and X.~Yang, ``Tiny imagenet visual recognition challenge,'' \emph{CS 231N}, vol.~7, no.~7, p.~3, 2015.

\bibitem{he2016deep}
K.~He, X.~Zhang, S.~Ren, and J.~Sun, ``Deep residual learning for image recognition,'' in \emph{Proceedings of the IEEE conference on computer vision and pattern recognition}, 2016, pp. 770--778.

\bibitem{zagoruyko2016wide}
S.~Zagoruyko and N.~Komodakis, ``Wide residual networks,'' \emph{arXiv preprint arXiv:1605.07146}, 2016.

\bibitem{xie2017aggregated}
S.~Xie, R.~Girshick, P.~Doll{\'a}r, Z.~Tu, and K.~He, ``Aggregated residual transformations for deep neural networks,'' in \emph{Proceedings of the IEEE conference on computer vision and pattern recognition}, 2017, pp. 1492--1500.

\bibitem{simonyan2014very}
K.~Simonyan and A.~Zisserman, ``Very deep convolutional networks for large-scale image recognition,'' \emph{arXiv preprint arXiv:1409.1556}, 2014.

\bibitem{sandler2018mobilenetv2}
M.~Sandler, A.~Howard, M.~Zhu, A.~Zhmoginov, and L.-C. Chen, ``Mobilenetv2: Inverted residuals and linear bottlenecks,'' in \emph{Proceedings of the IEEE conference on computer vision and pattern recognition}, 2018, pp. 4510--4520.

\bibitem{ma2018shufflenet}
N.~Ma, X.~Zhang, H.-T. Zheng, and J.~Sun, ``Shufflenet v2: Practical guidelines for efficient cnn architecture design,'' in \emph{Proceedings of the European conference on computer vision (ECCV)}, 2018, pp. 116--131.

\bibitem{zhang2018shufflenet}
X.~Zhang, X.~Zhou, M.~Lin, and J.~Sun, ``Shufflenet: An extremely efficient convolutional neural network for mobile devices,'' in \emph{Proceedings of the IEEE conference on computer vision and pattern recognition}, 2018, pp. 6848--6856.

\bibitem{peng2019correlation}
B.~Peng, X.~Jin, J.~Liu, D.~Li, Y.~Wu, Y.~Liu, S.~Zhou, and Z.~Zhang, ``Correlation congruence for knowledge distillation,'' in \emph{Proceedings of the IEEE/CVF International Conference on Computer Vision}, 2019, pp. 5007--5016.

\bibitem{9827964}
C.~Yang, Z.~An, L.~Cai, and Y.~Xu, ``Knowledge distillation using hierarchical self-supervision augmented distribution,'' \emph{IEEE Transactions on Neural Networks and Learning Systems}, vol.~35, no.~2, pp. 2094--2108, 2024.

\bibitem{song2023multi}
K.~Song, J.~Xie, S.~Zhang, and Z.~Luo, ``Multi-mode online knowledge distillation for self-supervised visual representation learning,'' in \emph{Proceedings of the IEEE/CVF Conference on Computer Vision and Pattern Recognition}, 2023, pp. 11\,848--11\,857.

\bibitem{Chen2020CrossLayerDW}
\BIBentryALTinterwordspacing
D.~Chen, J.-P. Mei, Y.~Zhang, C.~Wang, Z.~Wang, Y.~Feng, and C.~Chen, ``Cross-layer distillation with semantic calibration,'' in \emph{AAAI Conference on Artificial Intelligence}, 2020. [Online]. Available: \url{https://api.semanticscholar.org/CorpusID:227335337}
\BIBentrySTDinterwordspacing

\bibitem{szegedy2016rethinking}
C.~Szegedy, V.~Vanhoucke, S.~Ioffe, J.~Shlens, and Z.~Wojna, ``Rethinking the inception architecture for computer vision,'' in \emph{Proceedings of the IEEE conference on computer vision and pattern recognition}, 2016, pp. 2818--2826.

\bibitem{yun2019cutmix}
S.~Yun, D.~Han, S.~J. Oh, S.~Chun, J.~Choe, and Y.~Yoo, ``Cutmix: Regularization strategy to train strong classifiers with localizable features,'' in \emph{Proceedings of the IEEE/CVF international conference on computer vision}, 2019, pp. 6023--6032.

\end{thebibliography}

\end{document}